\documentclass[11pt,a4paper]{article}

\usepackage{amsmath, amssymb, amsthm}
\usepackage{microtype}
\usepackage{graphicx}
\usepackage{booktabs}
\usepackage{float}
\usepackage{enumitem}
\usepackage{hyperref}
\usepackage{lineno}
\usepackage[numbers,sort&compress]{natbib}
\usepackage{multirow}
\usepackage{xcolor}
\usepackage{ulem}
\usepackage{array}
\definecolor{brokenrefcolour}{RGB}{255,140,0}

\definecolor{trimcolour}{RGB}{220,90,0}

\definecolor{movedcolour}{RGB}{0,128,128}

\definecolor{revcolour}{RGB}{200,0,0}

\definecolor{updcolour}{RGB}{255,140,0}

\theoremstyle{plain}

\theoremstyle{definition}
\newtheorem{definition}{Definition}[section]

\theoremstyle{remark}
\newtheorem{remark}{Remark}[section]

\newcommand{\loss}{\operatorname{loss}}
\newcommand{\R}{\mathbb{R}}
\newcommand{\norm}[1]{\left\|#1\right\|}

\begin{document}

\noindent\rule{\textwidth}{2pt}

\vspace{0.3em}
\begin{center}
    {\LARGE \bfseries Identifying Representational Biases in Datasets Using PCA:\\
    A Max-Disparity Partition Framework}
\end{center}
\vspace{0.3em}

\noindent\rule{\textwidth}{0.4pt}

\vspace{1.5em}

\begin{center}
  \begin{minipage}[t]{0.4\textwidth}
    \centering
    \textbf{Arjun KM} \\
    Department of Management Studies \\
    Indian Institute of Science, Bangalore \\
    \small \texttt{arjunkm14@gmail.com}
  \end{minipage}%
  \hspace{4em}
  \begin{minipage}[t]{0.4\textwidth}
    \centering
    \textbf{Dr.\ Shashi Jain} \\
    Department of Management Studies \\
    Indian Institute of Science, Bangalore \\
    \small \texttt{shashijain@iisc.ac.in}
  \end{minipage}
\end{center}

\vspace{1.5em}

\begin{center}
August 2026
\end{center}

\vspace{1em}
%
\begin{abstract}
Principal Component Analysis (PCA) minimises aggregate reconstruction error,
which can inadvertently represent majority subgroups with substantially higher fidelity
than minority subgroups. Fairness-aware extensions of PCA correct this disparity but
require group labels as input. We address the logically prior question: given only a
data matrix, which binary partition of the data suffers the greatest representational
disparity under a shared PCA projection? We formalise this as the max-disparity
partition problem and propose a greedy local-search algorithm, grounded in the
Fiduccia-Mattheyses bipartitioning framework, that discovers the disparity-maximising
partition without any predefined group labels. Two benchmark algorithms a)
fixed-projection sorting baseline and b) simulated-annealing variant confirm that the
greedy solution is empirically near-optimal. Having identified the partition, we
attribute the disparity to specific features via PCA loading scores and association
rule mining, enabling a practitioner to assess whether the disadvantaged group
corresponds to a human-meaningful minority. On the Predict Students' Dropout and Academic Success dataset, representational disparity is driven predominantly by institutional and programmatic proxies for socioeconomic disadvantage, with gender emerging as a secondary but consistent contributor within the disadvantaged group. The discovered partition is then passed directly to Fair PCA~\citep{samadi2018fairpca}, completing a
detect-explain-mitigate pipeline.

\end{abstract}

\newpage


\section{Introduction}

Machine learning and artificial intelligence now automate consequential decisions in areas of healthcare, finance, education, and hiring. Because these systems learn from historical data, they risk encoding and amplifying the structural inequalities present in that data~\citep{mehrabi2021survey}. The fairness literature has responded with a rich body of pre-processing, in-processing, and post-processing techniques that constrain model behaviour with respect to known sensitive attributes such as gender or race. These techniques are well-developed for supervised settings, but the fairness implications of \emph{unsupervised} preprocessing steps,which precede any modelling task,remain far less understood.

Principal Component Analysis (PCA)~\citep{jolliffe1986pca} is perhaps the most widely used of those preprocessing steps. PCA finds an orthogonal basis for a low-dimensional subspace that maximises the variance of projected data, or equivalently minimises the Frobenius-norm reconstruction error over the entire dataset. This aggregate optimality conceals a structural asymmetry: the shared projection is computed from the joint covariance of all data points, so groups whose covariance structure differs from the majority will suffer disproportionately high reconstruction loss. The work of \citet{samadi2018fairpca} made this precise. They showed empirically that standard PCA incurs substantially higher average reconstruction error for women than for men on the Labeled Faces in the Wild dataset, even when male and female faces are sampled equiprobably,and for lower- versus higher-educated individuals on a credit dataset. Their \emph{Fair PCA} formulation corrects this by finding a $d$-dimensional projection that minimises the worst-case \emph{reconstruction loss} (the additional error above each group's own optimal projection) across predefined groups. A key theoretical result is that the optimal fair projection can be found in polynomial time and requires at most one extra dimension beyond the target rank~$d$.

The Fair PCA framework is powerful but rests on a critical assumption: the sensitive group labels are \emph{known in advance}. In practice this assumption is often untenable. Sensitive attributes may be legally prohibited from collection, may have been redacted, or may not manifest as any single explicit variable. More subtly, bias may arise not from a directly measured demographic feature but from \emph{proxy variables} or complex feature interactions that encode group membership indirectly. A system that asks ``how do we equalise reconstruction loss across groups?'' cannot be applied until the groups have been identified. The logically prior question is: \emph{given only the data matrix, which partition of the data suffers the greatest representational disparity under a shared PCA projection, and what drives that disparity?}

This paper answers that question. We introduce a two-stage framework that (i) discovers the binary partition that maximises reconstruction-loss disparity without any prior group label, and (ii) explains the discovered partition by attributing it to specific features, enabling a practitioner to assess whether the structurally disadvantaged group corresponds to a human-meaningful minority. Together, these two stages constitute a \emph{fairness audit}: a diagnostic that can precede, motivate, and guide the application of remedies such as Fair PCA.

\paragraph{The max-disparity partition problem.} Formally, given a data matrix $M \in \R^{m \times n}$ and a target rank $d$, we seek a binary partition $(A, B)$ of the $m$ rows that solves
\begin{equation}
  \max_{\substack{A \cup B = [m] \\ A \cap B = \emptyset}}
  \left|
    \frac{1}{|A|}\loss(A) -
    \frac{1}{|B|}\loss(B)
  \right|,
  \label{eq:max-disp}
\end{equation}
where, for each group $G \in \{A, B\}$,
\[
  \loss(G) \;=\; \norm{G - GP}_F^2 - \norm{G - \hat{G}}_F^2.
\]
Here $P = VV^\top$ is the rank-$d$ projection matrix obtained from the \emph{joint} PCA of $A \cup B$, so $GP$ is the reconstruction of group $G$ under the shared subspace; and $\hat{G}$ is the optimal rank-$d$ approximation of $G$ from PCA fitted on $G$ alone. Thus $\loss(G)$ measures the \emph{additional} reconstruction error that group $G$ incurs because it must share a projection with the other group, rather than using its own best subspace~\citep{samadi2018fairpca}. Problem~\eqref{eq:max-disp} is the \emph{dual} of Fair PCA: Fair PCA minimises the maximum loss given the groups; we maximise the difference in losses to discover the groups.

This optimisation is NP-hard in general: the projection $P$ depends on the partition (through the joint PCA of $A \cup B$), and the partition that maximises disparity cannot be read off from any simple closed-form expression. We are not aware of any analytical solution to the joint problem. We therefore pursue practical heuristic algorithms. Our primary algorithm is an iterative greedy procedure inspired by the Fiduccia-Mattheyses (FM) variant of the Kernighan-Lin (KL) local-search family for graph bipartitioning~\citep{kernighan1970efficient,fiduccia1982linear}. These algorithms are the standard practical tool for combinatorial partition problems: they converge quickly to a local optimum and, when paired with multiple random initialisations, reliably surface the dominant basin of the objective landscape. We complement the greedy search with two additional algorithms:a fixed-projection sorting baseline and a simulated-annealing variant and present a comparative analysis of all three approaches to determine which performs best on the disparity objective.

\paragraph{Interpretation and the bridge to human-meaningful fairness.} A partition of data points is only actionable if it can be explained. Once the max-disparity partition $(A, B)$ has been found, we apply two complementary attribution methods to identify the features that characterise and separate the two groups. The first uses PCA loading vectors to compute per-feature importance scores for each group separately; features that are both globally important and \emph{differentially} important between the two groups are flagged as primary drivers. The second employs association rule mining to capture higher-order feature interactions that distinguish the groups~\citep{agrawal1993mining}. The convergence of findings across these linear and combinatorial perspectives strengthens confidence in the identified drivers.

The interpretive step closes the loop between the mathematical disparity uncovered in Phase~1 and its societal significance. If the features that emerge as key drivers are known sensitive attributes (gender, age, socioeconomic status) or institutional proxies for them, such as daytime versus evening attendance as a proxy for working-student status, a practitioner has grounds to conclude that the discovered disparity is a fairness concern in the human sense, not merely a statistical artefact. Conversely, if the drivers are purely technical (measurement scale or data-collection differences), the disparity may have a different remediation path. This diagnostic capacity distinguishes our approach from methods that either assume group labels are available or that correct for bias without explaining its structural origin.

\paragraph{Mitigation.} Having identified which partition suffers disproportionate reconstruction loss and why, we demonstrate how the Fair PCA algorithm of \citet{samadi2018fairpca} can be applied as a principled mitigation step, using the discovered partition as the group definition that Fair PCA requires. This closes the pipeline: discover the disparity, explain it, then redress it.

\subsection*{Contributions and Related Work}

The fairness-in-ML literature has largely focused on supervised tasks-
classification
\citep{agarwal2019fair,aghaei2019learning,jiang2019wasserstein} and
ranking~\citep{kleinberg2017inherent},with fairness enforced through constraints on
statistical parity or equalised odds. A parallel stream targets unsupervised
representation learning. Several methods learn representations that are conditionally
independent of sensitive attributes~\citep{zemel2013learning,madras2018learning}, while
\citet{olfat2018spectral} formulate a convex programme that reduces the projected
data's dependence on a sensitive attribute. \citet{samadi2018fairpca} take a different
approach: rather than hiding the sensitive attribute, they ensure that each group's
reconstruction loss is close to its individual optimum, finding a $(d{+}1)$-dimensional
solution via a semidefinite programme. Spectral clustering~\citep{vonluxburg2007spectral}
and related unsupervised partition methods optimise geometric criteria unrelated to
reconstruction-loss disparity. All of the above methods take group membership as a
given input.

Our first contribution is to formalise the \emph{max-disparity partition problem}
(Definition~\ref{def:max-disp}), the dual of Fair PCA, which removes the requirement
for predefined group labels entirely. To our knowledge this problem has not previously
been posed in the PCA reconstruction-loss context.

Our second contribution is an efficient algorithm for this problem. The
Kernighan-Lin~\citep{kernighan1970efficient} and
Fiduccia-Mattheyses~\citep{fiduccia1982linear} algorithms are the canonical
local-search methods for graph bipartitioning; our greedy allocation-reallocation
procedure is a direct FM analogue with the edge-cut objective replaced by
$\Delta(A,B)$. We establish the term-by-term correspondence and show that FM's finite
convergence and local-optimality guarantees carry over. We complement this with a
fixed-projection sorting baseline, which is the exact solution to the inner sub-problem when the projection is held fixed, and a simulated-annealing
variant~\citep{kirkpatrick1983optimization} that quantifies the local-opti
mality gap
of the greedy solution.

Our third contribution is an attribution framework that connects the discovered
partition to human-interpretable features. PCA-based feature loading scores and
Apriori association rule mining~\citep{agrawal1993mining} are applied separately to
each group; the convergence of findings across these two independent methods
identifies robust structural drivers of the disparity. This attribution step is what
allows the mathematical quantity $\Delta(A,B)$ to be translated into a fairness
concern in the human sense.

Finally, we demonstrate a complete detect--explain--mitigate pipeline by passing the
discovered partition to Fair PCA~\citep{samadi2018fairpca}, showing that the
unsupervised audit directly enables the supervised correction.


\section{Problem Formulation}

We are given an $n$-dimensional dataset represented as the rows of a matrix $M \in \R^{m \times n}$. All features are normalised to $[0,1]$ prior to analysis; no feature is designated as sensitive, and no group label is assumed to be available.

\begin{definition}[PCA problem]
\label{def:pca}
Given $M \in \R^{m \times n}$, find $\hat{M} \in \R^{m \times n}$ of rank at most
$d \le n$ that minimises $\norm{M - \hat{M}}_F$. The solution is $\hat{M} = MWW^\top$
where $W \in \R^{n \times d}$ has orthonormal columns equal to the top-$d$ eigenvectors
of $M^\top M$, and $P = WW^\top$ is the corresponding projection matrix.
\end{definition}

\begin{definition}[Reconstruction loss \citep{samadi2018fairpca}]
\label{def:loss}
Let $G \in \R^{a \times n}$ be a group of data points with optimal rank-$d$
approximation $\hat{G}$. Given a projection $P$ obtained from PCA fitted on a
superset of $G$, the reconstruction loss of $G$ under $P$ is
\[
  \loss(G, P) \;=\; \norm{G - GP}_F^2 - \norm{G - \hat{G}}_F^2.
\]
This is the additional Frobenius error that $G$ incurs from using the shared
projection $P$ rather than its own optimal subspace.
\end{definition}

\begin{definition}[Max-disparity partition problem]
\label{def:max-disp}
Given $M \in \R^{m \times n}$, target rank $d$, and minimum group size $r \ge 1$, find
a binary partition $(A, B)$ with $A \cup B = [m]$, $A \cap B = \emptyset$,
$|A|, |B| \ge r$, that maximises
\[
  \Delta(A,B) \;:=\; \left|
    \frac{1}{|A|}\loss(A, P) - \frac{1}{|B|}\loss(B, P)
  \right|,
\]
where $P = VV^\top$ is the rank-$d$ projection obtained from the joint PCA of
$A \cup B$.
\end{definition}

\begin{remark}
Definition~\ref{def:max-disp} is the \emph{dual} of Fair PCA~\citep{samadi2018fairpca}:
Fair PCA minimises $\max\!\left\{\frac{1}{|A|}\loss(A,P),
\frac{1}{|B|}\loss(B,P)\right\}$ over projections $P$ for fixed groups; we maximise
the difference over partitions $(A,B)$ with $P$ co-determined by the partition.
\end{remark}

\begin{remark}[Fixed-projection inner optimum]
\label{rem:sort}
For a \emph{fixed} projection matrix $P$, the per-point reconstruction error
$e_i = \norm{x_i - x_i P}^2$ is well-defined for each row $x_i$. For fixed $P$, the
max-disparity partition is solved exactly by sorting all points by $e_i$ and assigning
the top half to the high-loss group. The difficulty of the full problem arises because
$P$ depends on the partition.
\end{remark}

\noindent We refer to the group bearing the higher average reconstruction loss as the \emph{minority group} and the other as the \emph{majority group}.

\subsection*{Variable Contribution Assessment}

To attribute the discovered disparity to specific features, we use two complementary
methods. The first is a PCA-based score: let $V \in \R^{n \times d}$ be the loading matrix
from PCA of group $G$, with normalised component contribution $C_{ij} = v_{ij}^2 / \sum_k
v_{kj}^2$. The overall feature score $s(f_i) = \sum_{j=1}^{d} \lambda_j C_{ij}$ weights
each feature by the variance it explains in each component. Scores are computed separately
for the full dataset, the minority group, and the majority group; the disparity score
$\Delta_s(f_i) = |\bar{s}_{\text{maj}}(f_i) - \bar{s}_{\text{min}}(f_i)|$ measures how
asymmetrically feature $f_i$ contributes across groups.

The second method is association rule mining. We apply the Apriori
algorithm~\citep{agrawal1993mining} to each group separately, retaining rules
$X \Rightarrow Y$ by support, confidence, and lift. Features that appear consistently
in high-lift rules for one group but not the other are flagged as interaction-level
drivers that linear attribution cannot expose.


\section{Methodology}

\subsection{Determining the Projection Dimensionality}

For a given dataset $M \in \R^{m \times n}$, we first select the projection dimensionality $d$ using a scree plot of cumulative explained variance. The $j$-th explained variance ratio is $\mathrm{EVR}_j = \lambda_j / \sum_i \lambda_i$, and $d$ is chosen as the smallest value such that $\sum_{j=1}^{d} \mathrm{EVR}_j$ exceeds a threshold (typically 80\%). This value is held fixed across all downstream analyses.

\subsection{Partition Algorithms}

\subsubsection{Algorithm~1: Greedy Allocation and Reallocation (Primary)}

\paragraph{Connection to Kernighan--Lin and Fiduccia--Mattheyses.} 

\begin{sloppypar}
The Kernighan--Lin (KL) algorithm~\citep{kernighan1970efficient} and its Fiduccia--Mattheyses (FM) refinement~\citep{fiduccia1982linear} are the canonical local-search methods for graph bipartitioning. Both maintain a binary partition $(A, B)$ and improve it by moving individual elements from one side to the other. Table~\ref{tab:klfm} maps the KL/FM framework onto our algorithm term by term.
\end{sloppypar}

\begin{table}[H]
\centering
\small
\resizebox{\textwidth}{!}{%
\begin{tabular}{@{}lll@{}}
\toprule
\textbf{KL / FM concept} & \textbf{KL / FM instantiation} & \textbf{Our instantiation} \\
\midrule
Ground set         & Vertices $V$ of a graph
                   & Data points $\{x_1,\ldots,x_m\}$ \\
Partition          & $(A, B)$, $A \cup B = V$
                   & $(A, B)$, $A \cup B = [m]$ \\
Objective          & Edge-cut $\mathrm{cut}(A,B)$
                   & Loss disparity $\Delta(A,B)$ \\
Optimisation dir.  & Minimise cut
                   & Maximise $\Delta$ \\
Elementary move    & Move one vertex $v$ from $A$ to $B$
                   & Move one point $x_i$ from $A$ to $B$ \\
Move acceptance    & Accept if new cut $<$ current cut
                   & Accept if new $\Delta >$ current $\Delta$ \\
Objective update   & $O(|E|)$ incremental (FM)
                   & Shared $P$ fixed per epoch; recompute group PCAs $P_A$, $P_B$ \\
Balance constraint & $\bigl||A| - |B|\bigr| \le 1$
                   & $|A|, |B| \ge r = 5n$ \\
Termination        & No improving move exists
                   & $\Delta_t$ converges or $T$ epochs reached \\
\bottomrule
\end{tabular}
}
\caption{Term-by-term correspondence between FM graph bipartitioning and our
         greedy reallocation algorithm (Stage~2).}
\label{tab:klfm}
\end{table}

\noindent Two properties carry over directly from FM to our setting.
\begin{itemize}[leftmargin=1.5em]
  \item \textbf{Finite convergence.} The objective $\Delta_t$ is non-negative and
  bounded above (by the total reconstruction error of $M$). Each accepted move
  strictly increases $\Delta_t$, and the number of distinct partitions is finite.
  Therefore the reallocation loop terminates in finite steps at a local optimum,
  exactly as FM terminates at a local minimum of the edge-cut.

  \item \textbf{Local-optimality guarantee.} At termination, no single-point transfer
  can increase $\Delta$. This is the FM definition of a locally optimal partition,
  transplanted verbatim to our objective.
\end{itemize}

\noindent One property does \emph{not} carry over. In FM, removing a vertex $v$ from a graph partition changes the cut by a quantity computable in $O(\deg(v))$ time, so each move evaluation is cheap. In our setting the situation is more nuanced. Because $A \cup B = M$ at every epoch (every point is assigned to exactly one group), the shared projection $P^{(t)}$ is simply the rank-$d$ PCA of the full dataset $M$ and is \emph{fixed} throughout the epoch --- it does not change when a point is moved. What does change are the group-specific optimal projections $P_A^{(t)}$ and $P_B^{(t)}$, which are PCA fitted on $A^{(t)}$ and $B^{(t)}$ individually. Moving one point $x_i$ from $A$ to $B$ modifies both group covariance matrices by a rank-one update, so in principle the updated group PCAs could be obtained via rank-one SVD update formulae rather than a full recomputation. However, since both group sizes are $O(m)$, the practical saving is modest, and in our implementation we recompute both group PCAs from scratch at each evaluation. The per-epoch cost is therefore $O(m \cdot \mathrm{PCA}(m,n,d))$, where $\mathrm{PCA}(m,n,d)$ denotes the cost of a rank-$d$ SVD.

\paragraph{Stage~1 -Greedy allocation.} Two seed sets $A^{(0)}$ and $B^{(0)}$ are initialised by drawing $5n$ points each at random without replacement. Each remaining point $x \in M \setminus (A^{(0)} \cup B^{(0)})$ is then assigned as follows. For each candidate group $G \in \{A, B\}$:
\begin{enumerate}[leftmargin=2em,label=\arabic*.]
  \item Fit a rank-$d$ PCA to $(A \cup B \cup \{x\})$ and form the shared projection
        $P = VV^\top$.
  \item Fit rank-$d$ PCAs to $(A \cup \{x\})$ and $(B \cup \{x\})$ separately to
        obtain group-optimal projections $P_A = V_A V_A^\top$ and
        $P_B = V_B V_B^\top$.
  \item Compute the average reconstruction loss for each group under the shared projection:
        \[
          \loss(A, P) = \tfrac{1}{|A|}\norm{AP - AP_A}_F^2, \qquad
          \loss(B, P) = \tfrac{1}{|B|}\norm{BP - BP_B}_F^2.
        \]
\end{enumerate}
Assign $x$ to whichever group produces the larger reconstruction loss upon inclusion:
\[
  x \;\rightarrow\;
  \begin{cases}
    A & \text{if } \loss(A, P) > \loss(B, P), \\
    B & \text{otherwise.}
  \end{cases}
\]
\paragraph{Stage~2 - Greedy reallocation.} Starting from $(A^{(0)}, B^{(0)})$, run $T$ refinement epochs. At epoch $t$, compute:
\begin{align*}
  P^{(t)} &= V^{(t)}{V^{(t)}}^\top
            \quad\text{(joint PCA of } A^{(t)} \cup B^{(t)}\text{)}, \\
  \Delta_t &= \Bigl|\tfrac{1}{|A^{(t)}|}\loss(A^{(t)}, P^{(t)}) - \tfrac{1}{|B^{(t)}|}\loss(B^{(t)}, P^{(t)})\Bigr|,
\end{align*}
where $\loss(G^{(t)}, P^{(t)}) = \norm{G^{(t)} P^{(t)} - G^{(t)} P_G^{(t)}}_F^2$ and $P_G^{(t)}$ is the rank-$d$ projection from PCA fitted on $G^{(t)}$ alone. For each point $x_i$, evaluate transferring it to the opposite group. Accept the transfer if and only if the resulting disparity $\Delta_t' > \Delta_t$. Enforce minimum group size $r = 5n$ throughout. Output the partition at $t^* = \arg\max_t \Delta_t$.

\subsubsection{Algorithm~2: Fixed-Projection Sorting Baseline}

Remark~\ref{rem:sort} shows that for a fixed projection $P$ the optimal partition is a sort by individual errors. This motivates a simple, single-pass benchmark that requires only one PCA fit.

\paragraph{Algorithm.}
\begin{enumerate}[leftmargin=2em,label=\arabic*.]
  \item Run standard PCA on $M$ to obtain the global rank-$d$ projection
        $P_0 = V_0 V_0^\top$.
  \item Compute per-point reconstruction errors:
        \[
          e_i = \norm{x_i - x_i P_0}^2, \quad i = 1,\ldots,m.
        \]
  \item Sort all points by $e_i$ in descending order. Assign the top
        $\lfloor m/2 \rfloor$ points to the high-loss group $A$ and the remainder to
        $B$, subject to minimum group size $r$.
\end{enumerate}

\paragraph{Rationale.} This algorithm solves the inner sub-problem of Definition~\ref{def:max-disp} optimally for the global PCA projection $P_0$. It serves two purposes: (a) a fast lower bound on the max-disparity objective requiring no iteration, and (b) a comparison point to assess how much the co-evolution of partition and projection matters in practice. If the sorting baseline and the greedy algorithm produce similar partitions (high Jaccard similarity and close $\Delta$ values), the simpler baseline suffices; if they diverge substantially, the iterative refinement of Algorithm~1 is necessary.

\subsubsection{Algorithm~3: Simulated Annealing}

Neither Algorithm~1 nor Algorithm~2 guarantees proximity to the global optimum of Definition~\ref{def:max-disp}. Simulated annealing (SA)~\citep{kirkpatrick1983optimization} is the standard metaheuristic for escaping local optima in combinatorial problems. We implement it here as a benchmark to bound the local-optimality gap of the greedy method.

\paragraph{Algorithm.}
\begin{enumerate}[leftmargin=2em,label=\arabic*.]
  \item \textit{Initialisation.} Start from the greedy partition $(A^*, B^*)$ produced
        by Algorithm~1 (warm start) prior to the Stage~2 reallocation
        refinement. Set initial temperature $T_0 = \Delta^*$ (the
        greedy $\Delta$ value) and cooling rate $\alpha \in (0,1)$.
  \item \textit{Proposal.} Uniformly sample a point $x_i$ and propose moving it to the
        opposite group, yielding $(A', B')$. If $|A'| < r$ or $|B'| < r$, resample.
  \item \textit{Evaluation.} Recompute the shared projection $P'$ from the joint PCA
        of $A' \cup B'$ and compute
        $\Delta' = {\left|\tfrac{1}{|A'|}\loss(A', P') - \tfrac{1}{|B'|}\loss(B', P')\right|}$.
  \item \textit{Acceptance.} Let $\Delta_{\mathrm{cur}}$ denote the disparity of the
        current partition. Accept the move with probability
        \[
          p_{\mathrm{accept}} =
          \begin{cases}
            1
              & \text{if } \Delta' > \Delta_{\mathrm{cur}}, \\[4pt]
            \exp\!\left(\dfrac{\Delta' - \Delta_{\mathrm{cur}}}{T_k}\right)
              & \text{otherwise.}
          \end{cases}
        \]
  \item \textit{Cooling.} Update $T_{k+1} = \alpha\, T_k$. Record the best partition
        seen across all steps.
  \item \textit{Termination.} Stop when $T_k < T_{\min}$ or after $K_{\max}$ steps.
\end{enumerate}

\paragraph{Hyperparameters.} We set $\alpha = 0.999$, $T_{\min} = 10^{-6}$, and $K_{\max} = 10{,}000$. The slow geometric cooling rate $\alpha = 0.999$ ensures that the temperature decays gradually over the full run rather than collapsing to the floor in the first few hundred steps, so that SA's exploratory behaviour is preserved across a substantial fraction of the iterations. Setting $T_0$ proportional to the warm-start disparity ensures that, at the outset, modest downhill moves are accepted with non-trivial probability, providing meaningful exploration without diverging wildly from the warm-start solution.

\paragraph{Role in the analysis.} SA answers the key validation question: does the greedy algorithm find a partition close to the global optimum? We compare SA and greedy on (a) final $\Delta$ value, (b)~Jaccard similarity between the two final partitions, and (c) feature-importance rankings. High agreement across all three confirms that the greedy local optimum is representative and that the downstream interpretive analysis is not biased by a suboptimal partition.

\subsection{Summary of the Three Algorithms}

Table~\ref{tab:algo-comparison} places the three approaches side by side.

\begin{table}[H]
\centering
\small
\resizebox{\textwidth}{!}{%
\begin{tabular}{@{}llll@{}}
\toprule
\textbf{Algorithm} & \textbf{Cost per run} & \textbf{Optimality} & \textbf{Role} \\
\midrule
Greedy alloc.\ + realloc.\ (Alg.~1)
  & $O(T \cdot m \cdot \mathrm{PCA})$
  & Local opt.\ (FM-style)
  & Primary method  \\[2pt]
Fixed-projection sort (Alg.~2)
  & $O(m\log m + \mathrm{PCA})$
  & Optimal for fixed $P_0$
  & Fast lower bound \\[2pt]
Simulated annealing (Alg.~3)
  & $O(K_{\max} \cdot \mathrm{PCA})$
  & Empirical near-global
  & Global-search benchmark \\
\bottomrule
\end{tabular}
}
\caption{Comparison of the three partition algorithms. PCA cost is
$O(\min(m,n)^2 \max(m,n))$ via SVD; $T$ = number of epochs;
$K_{\max}$ = number of SA steps.}
\label{tab:algo-comparison}
\end{table}


\section{Results and Analysis}

\subsection{Dataset and Preprocessing}

We validate the framework on the publicly available Predict Students' Dropout and Academic Success dataset (Realinho et al., 2022), released in December 2021 to support research on student attrition in higher education. It comprises 4,425 student records described by 36 explanatory variables spanning demographic profile (e.g., age, gender, nationality), academic history (e.g., previous qualification, admission grade), enrollment characteristics (e.g., application order, daytime or evening attendance), and institutional attributes (e.g., curricular structure, financial status).
To ensure consistent benchmarking across multiple experimental runs under reasonable computational cost, a stratified random sample of 2,000 observations is drawn from the full dataset. Categorical variables are then target-encoded, replacing each category with its observed average graduation rate, and all features are normalised to the common range $[0,1]$ to permit equitable comparison across heterogeneous scales and distributions.

\vspace{1em}

\subsection{PCA and Dimensionality Selection}

\begin{figure}[H]
    \centering
    \includegraphics[width=0.75\textwidth]{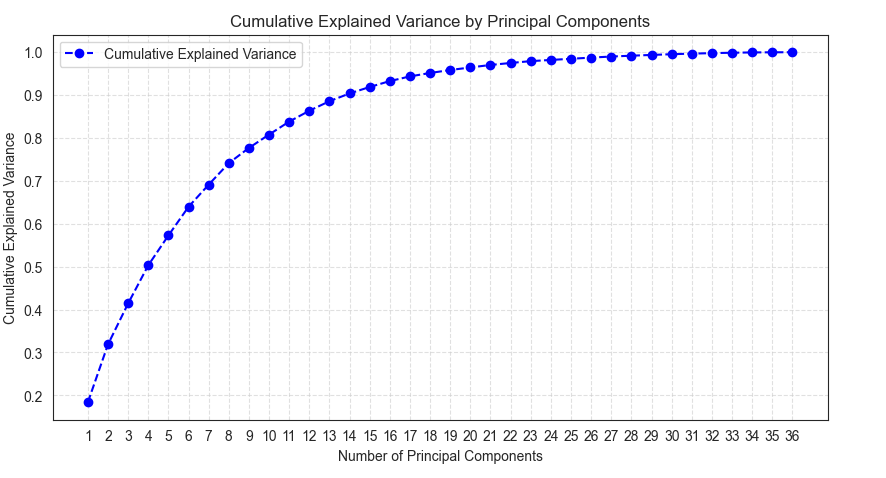}
    \caption{Cumulative variance explained by principal components.
    The first nine components capture nearly 80\% of the dataset's total variance.}
    \label{fig:explained_variance}
\end{figure}

We apply PCA to the normalised matrix $M \in \R^{2000 \times 36}$ and retain
components up to 80\% cumulative explained variance. As shown in
Figure~\ref{fig:explained_variance}, the first nine components suffice; we fix
$d = 9$ for all downstream analyses.

\noindent

\subsection{Reconstruction Loss Disparity}

We now report the empirical behaviour of our greedy algorithm(greedy allocation- reallocation) on the preprocessed dataset.

\begin{figure}[H]
    \centering
    \includegraphics[width=0.75\textwidth]{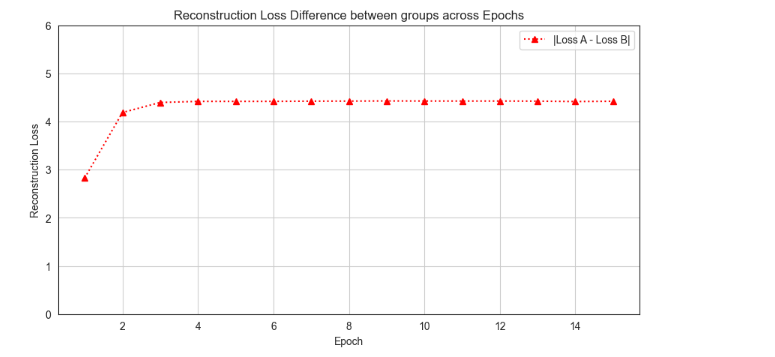}
    \caption { Reconstruction loss difference between Groups A and B across 15 epochs. The absolute loss difference is tracked at each iteration to monitor the emergence of representational disparity.}
    \label{fig:reconstruction_loss}
\end{figure}

The effectiveness of the iterative reallocation procedure is evaluated by tracking the evolution of reconstruction loss difference between groups across epochs. For this study, we fix the total number of epochs to $T = 15$, to balance convergence with computational efficiency. As shown in Figure~\ref{fig:reconstruction_loss}, the absolute reconstruction loss gap $\Delta_t$ steadily increases during the early epochs, indicating that the algorithm progressively refines group boundaries to maximize divergence in representational quality. The separation trend stabilizes after certain epochs, beyond which further iterations yield marginal improvements.

\subsection{Robustness of Groupings}

To assess stability and reproducibility of the greedy partition, we run the two-stage algorithm across 30 independent trials, each initialised with a distinct random seed.The choice of 30 runs aligns with the law of large numbers and standard statistical practice, ensuring that the observed stability metrics are not artifacts of sampling variability but reflect the algorithm’s consistent convergence behaviour.

\vspace{1em}

\begin{figure}[H]
    \centering
    \includegraphics[width=0.75\textwidth]{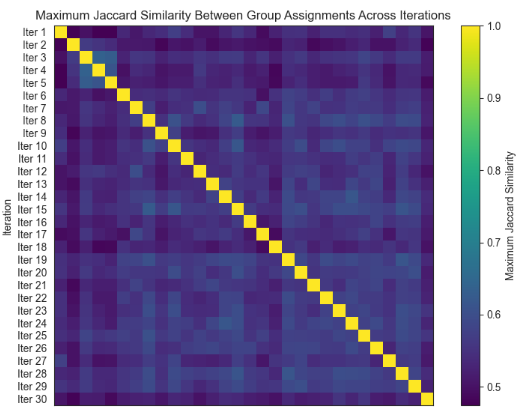}
    \caption  
    {Distribution of Jaccard similarity scores across runs
    initialised with different random seeds}
    \label{fig:jaccard_similarity}
\end{figure}

\textbf{Cross-run Jaccard similarity.} Each run produces a final partition $(A^{(i)}, B^{(i)})$ for $i = 1,\ldots,30$. We compute the maximum Jaccard similarity between every pair of runs (accounting for the arbitrary labelling of $A$ versus $B$) and visualise the resulting $30 \times 30$ similarity matrix in Figure~\ref{fig:jaccard_similarity}. Our results show that the pairwise Jaccard similarity across runs predominantly fell within the range of 0.53 to 0.64 with a central tendency of 0.55, indicating that despite randomness in initial group seeding, the algorithm converges toward a broadly consistent partitioning regime.

\begin{figure}[H]
    \centering
    \includegraphics[width=0.75\textwidth]{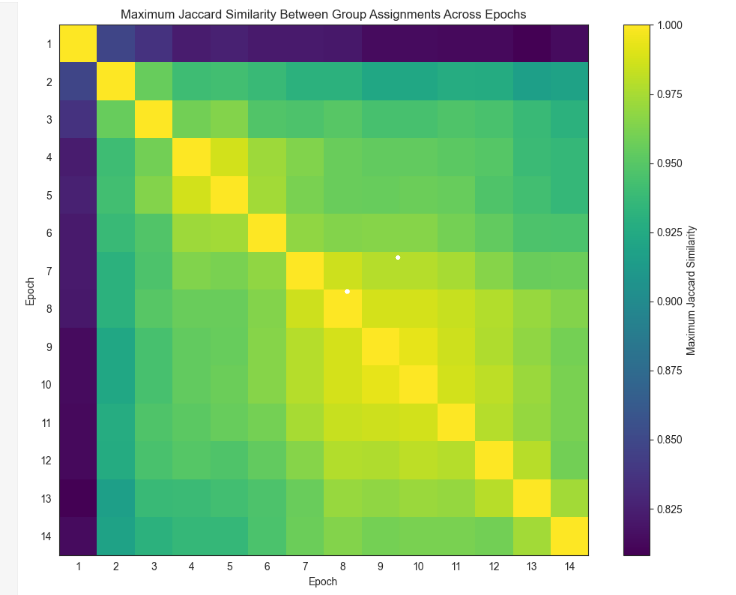}
    \caption {Intra-run Jaccard similarity across epochs.}
    \label{fig:reallocation_stability}
\end{figure}

\noindent \textbf{Intra-epoch stability.} Within each run, we track the Jaccard similarity between consecutive epochs across the 15 reallocation steps. The similarity remains consistently above 0.99 between adjacent epochs, indicating that fewer than 1\% of instances migrate between groups during any single reallocation step. Furthermore, the similarity remains above 0.80 between the first and final epochs, implying that more than 80\% of group memberships remain unchanged throughout the entire reallocation process (Figure~\ref{fig:reallocation_stability}).

\subsection{Comparative Analysis of the Three Proposed Algorithms}

The previous subsection established that the greedy procedure produces reproducible partitions across random initializations. An equally important question, which the Jaccard analysis does not address, is whether the resulting disparity is close to what a more global optimization of the same objective would achieve. Since the joint problem in Definition~\ref{def:max-disp} admits no analytical solution, we determine empirically which of the three proposed algorithms maximises the disparity $\Delta$ in practice. We therefore compare our greedy procedure against the two other proposed algorithms defined in the Methodology: Fixed Projection Sorting (FPS, Algorithm~2) and Simulated Annealing (SA, Algorithm~3).

All methods are evaluated on the same preprocessed dataset and under the same minimum group-size constraint $5n$. For FPS, both the equal-split variant ($|A|=|B|=m/2$) and the matched-split variant (sizes fixed to those of our greedy output) are reported. Simulated annealing was run for 10{,}000 single-point proposals with $\alpha = 0.999$, $T_{\min} = 10^{-6}$, and $T_0$ initialized proportionally to the greedy-filled initial disparity. Table~\ref{tab:benchmark_losses} reports the group-wise losses, the absolute disparity $\Delta = \big|\tfrac{1}{|A|}\operatorname{loss}_A - \tfrac{1}{|B|}\operatorname{loss}_B\big|$, and the resulting group sizes for each method.

\begin{table}[H]
\centering
\begin{tabular}{lrrrrr}
\hline
Method & $\operatorname{loss}_A$ & $\operatorname{loss}_B$ & $\Delta$ & $|A|$ & $|B|$ \\
\hline
Greedy  & 0.0625 & 5.4576 & \textbf{5.3951} & 1258 & 742 \\
FPS - equal split & 0.5022 & 0.4056 & 0.0966 & 1000 & 1000 \\
FPS - matched split & 0.4995 & 0.5296 & 0.0301 & 1258 & 742 \\
Simulated Annealing & 0.0270 & 4.6861 & 4.6591 & 1555 & 445 \\
\hline
\end{tabular}

\caption{Group-wise reconstruction losses, disparity $\Delta$, and final group sizes for the greedy procedure and the two alternative optimizers. The greedy procedure attains the largest $\Delta$ among all methods, including simulated annealing with 10{,}000 proposals.}
\label{tab:benchmark_losses}
\end{table}

Two observations emerge from Table~\ref{tab:benchmark_losses}. First, the greedy
procedure attains the largest disparity ($\Delta = 5.3951$) among all methods considered,
including simulated annealing, which is specifically designed to escape local optima
through stochastic exploration. 10{,}000 SA proposals reach a disparity of only
$\Delta = 4.6591$, below the greedy value. This is empirical evidence that our
greedy solution lies in a high-quality region of the optimisation landscape that a
properly tuned global search does not surpass.
Second, the  gap between the greedy disparity and the disparities attained by the two FPS variants reflects the coupled nature of the objective rather than a shortcoming of either method. Under FPS, $P$ is computed once from the full data and held fixed, so the resulting partition is optimal only under that neutral projection. Our procedure, by contrast, co-adapts the partition and the projection through iterative reallocation, which is precisely what allows $\Delta$ to grow substantially. The comparison therefore clarifies that the disparity reported in the previous subsection is attributable to the joint optimization of partition and subspace, and is not an artifact of the sorting rule alone.

\subsection{Variable Importance Assessment}
Having established the partition's existence and stability, we now turn to its interpretation: which features explain the observed representational disparity? We adopt a dual attribution strategy that combines a PCA-based feature scoring framework with association rule mining. The two perspectives are complementary: PCA loadings capture the linear contribution of each variable to the subspace structure, while rule mining surfaces non-linear and interaction-driven patterns that linear attribution cannot expose. Convergence between them strengthens the case that any feature flagged as a driver is genuinely so.

\subsubsection{PCA-Based Feature Importance Assessment}
We compute PCA-based feature scores across 30 independent runs of Algorithm~1,
each initialised with a distinct random seed, and aggregate the scores by averaging.
Scores are computed separately for the full dataset, the minority group, and the
majority group, following the formulation in Section~4. The absolute loss-difference
score, $\Delta_s(f_i)$, additionally quantifies the asymmetry of each feature's
contribution between the two groups.

Table~\ref{tab:topfeatures} summarises the top five features under each of the
four metrics; complete scores for all 36 variables are reported in
Table~\ref{tab:feature_scores_appendix} of Appendix~A.

\begin{table}[h]
\centering
\small
\resizebox{\textwidth}{!}{%
\begin{tabular}{p{3.4cm} p{3.4cm} p{3.4cm} p{3.4cm}}
\toprule
\textbf{Total Score} & \textbf{Minority Score} & \textbf{Majority Score} & \textbf{LossDiff Score} \\
\midrule
Daytime/evening attendance & Application order & Daytime/evening attendance & Daytime/evening attendance \\
Application order & Course & Application order & Course \\
Application mode & Gender & Application mode & Application order \\
GDP & Daytime/evening attendance & GDP & Gender \\
Course & GDP & Unemployment rate & Application mode \\
\bottomrule
\end{tabular}
}
\caption{Top Features across all iterations based on following Metric: Total Importance, Minority Score, Majority Score, and Loss Difference Score}
\label{tab:topfeatures}
\end{table}

Two variables stand out across every metric. \textbf{Daytime/evening attendance} and \textbf{Application order} dominate both the total importance ranking and the loss-difference ranking, identifying them as the principal axes along which the two groups separate. Beyond this shared leadership, the rankings reveal a clear asymmetry between groups. Within the minority partition, the leading contributors are programmatic and demographic such as \textbf{Application order}, \textbf{Course}, and \textbf{Gender} suggesting that the disadvantaged subgroup is characterised by the academic track a student enters and by demographic composition. Within the majority partition, by contrast, representational weight concentrates overwhelmingly in \textbf{Daytime/evening attendance}, reinforced by administrative and macroeconomic variables in application\textbf{ order}, \textbf{Application mode}, and \textbf{GDP}. The contrast indicates that representational disparity is not driven by a single dimension common to both groups, but by distinct mechanisms that map onto different institutional realities.

The loss-difference ranking sharpens this picture. \textbf{Daytime/evening attendance} exhibits by far the largest asymmetry between groups, followed by \textbf{Course} and \textbf{Gender}, confirming that these three features carry the bulk of the representational gap. The prominence of \textbf{Course} and \textbf{Gender} within the minority partition indicates that the disadvantaged group is delineated not by a single institutional axis but by a combination of programmatic placement and demographic membership.

At the opposite end of the ranking, \textbf{Nationality}, \textbf{Parental occupation}, and \textbf{Educational special needs} contribute negligibly under every metric. The disparity is therefore concentrated in a small number of programmatic, scheduling, and demographic variables, while the long tail of administrative and family-background attributes plays essentially no role in separating the two groups.

\subsubsection{Association Rule Mining and Interaction effects}
To complement the linear PCA-based attribution, we apply association rule mining to recover non-linear and interaction-driven patterns that linear loadings cannot capture. Each observation is encoded as a transaction of discretised attribute--value pairs, and we extract rules of the canonical form
\[
X \Rightarrow Y,
\]
where the consequent $Y$ is the group label and the antecedent $X$ is a conjunction of one or two feature--value pairs. Rules are mined for the minority and majority labels across the same 30 independent runs used for the PCA analysis. For each run we retain rules above a minimum confidence floor, rank the survivors by lift, and record the top antecedents per group. The frequency with which each variable surfaces as a top-ranked univariate antecedent across the 30 runs is summarised in Table~\ref{tab:rulefreq}.

\begin{table}[h]
\centering
\small
\resizebox{\textwidth}{!}{%
\begin{tabular}{lcc}
\toprule
\textbf{Variable} & \textbf{Minority Group (frequency)} & \textbf{Majority Group (frequency)} \\
\midrule
Daytime/evening attendance & 1/30 & 29/30 \\
Course & 29/30 & 13/30 \\
Application mode & 20/30 & 28/30 \\
Tuition fees up to date & 0/30 & 24/30 \\
Curricular units 1st sem (grade) & 23/30 & 29/30 \\
Inflation rate & 21/30 & 8/30 \\
GDP & 16/30 & 8/30 \\
Unemployment rate & 16/30 & 4/30 \\
Debtor & 7/30 & 2/30 \\
Gender & 1/30 & 2/30 \\
\bottomrule
\end{tabular}
}
\caption{Frequency of key variables appearing in high-lift univariate rules across 30 runs, separated by Minority and Majority groups.}
\label{tab:rulefreq}
\end{table}

The rule-mining results corroborate the PCA-based ranking on the variables that most sharply separate the two groups. \textbf{Daytime/evening attendance} is the dominant majority-group antecedent (29/30 runs) yet almost never characterises the minority (1/30), confirming the PCA finding that attendance timing anchors the majority partition. Conversely, \textbf{Course} is the leading minority-group antecedent (29/30) and considerably weaker in the majority (13/30), again matching the loading-based attribution in which Course was the principal minority driver. \textbf{Tuition fees up to date} surfaces exclusively in the majority group (24/30 versus 0/30), while the macroeconomic variables such as \textbf{Inflation rate}, \textbf{GDP}, and \textbf{Unemployment rate} are markedly more frequent in the minority partition, indicating that the disadvantaged subgroup is distinguished partly by sensitivity to economic context. \textbf{Application mode} and \textbf{Curricular units 1st sem (grade)} appear frequently in both groups and are therefore shared rather than discriminating. The convergence of two methodologically distinct attribution methods like linear PCA loadings and combinatorial rule mining on the same group assignment reinforces the structural origin of the observed disparity.

\paragraph{Joint Feature Contributions.} Univariate rules identify the strongest single antecedents but cannot capture interaction effects in which two features jointly delineate a subgroup. To recover these, we examine two-antecedent rules across the 30 runs, recording for each pair its frequency of appearance and the range of lift values it attains. The most frequent interaction pairs for each group are summarised in Table~\ref{tab:interactions}.

\begin{table}[h]
\centering
\small
\resizebox{\textwidth}{!}{%
\begin{tabular}{p{9cm}cc}
\toprule
\textbf{Interaction (antecedent pair)} & \textbf{Frequency (runs)} & \textbf{Lift range} \\
\midrule
\multicolumn{3}{l}{\textit{Minority group}} \\
Curricular units 2nd sem (grade) $\times$ Displaced & 21/30 & 1.23--1.54 \\
Course $\times$ Curricular units 1st sem (approved) & 21/30 & 1.25--1.46 \\
Course $\times$ Curricular units 2nd sem (grade) & 19/30 & 1.23--1.42 \\
Course $\times$ Mother's qualification & 14/30 & 1.21--1.36 \\
Curricular units 1st sem (grade) $\times$ Displaced & 11/30 & 1.23--1.45 \\
Curricular units 2nd sem (enrolled) $\times$ Gender & 11/30 & 1.24--1.33 \\
Course $\times$ Gender & 10/30 & 1.24--1.34 \\
\midrule
\multicolumn{3}{l}{\textit{Majority group}} \\
Curricular units 2nd sem (grade) $\times$ Tuition fees up to date & 22/30 & 1.33--1.61 \\
Curricular units 2nd sem (approved) $\times$ Tuition fees up to date & 21/30 & 1.36--1.62 \\
Curricular units 1st sem (approved) $\times$ Tuition fees up to date & 21/30 & 1.34--1.61 \\
Curricular units 2nd sem (grade) $\times$ Gender & 21/30 & 1.30--1.50 \\
Curricular units 1st sem (grade) $\times$ Gender & 20/30 & 1.31--1.51 \\
Application mode $\times$ Curricular units 2nd sem (grade) & 18/30 & 1.36--1.60 \\
Daytime/evening attendance $\times$ Gender & 16/30 & 1.33--1.49 \\
\bottomrule
\end{tabular}
}
\caption{Most frequent high-lift two-antecedent interaction effects across 30 experimental runs, with the range of lift values attained, separated by group.}
\label{tab:interactions}
\end{table}

The interaction analysis reinforces and extends the univariate picture. Within the minority group, \textbf{Course} recurs as the dominant interacting variable, pairing with academic-performance indicators (Curricular units 1st/2nd sem) and with demographic covariates (Mother's qualification, Gender); the recurring \textbf{Displaced}-based pairs further indicate that displacement combines with academic standing to delineate the disadvantaged subgroup. Within the majority group, \textbf{Tuition fees up to date} is the dominant interacting variable, pairing repeatedly with curricular-performance indicators at the highest lift values observed (up to 1.62), and \textbf{Daytime/evening attendance} reappears in combination with Gender (16/30), consistent with its role as the leading majority antecedent in the univariate analysis. Two observations stand out. First, the majority-group interactions attain systematically higher lift (commonly 1.4--1.6) than the minority-group interactions (commonly 1.2--1.4), indicating tighter, more concentrated patterns in the majority partition. Second, the joint analysis surfaces structure that single-feature attribution cannot: pairings such as Course $\times$ academic performance and Tuition status $\times$ curricular units are genuine interaction effects, not reducible to the marginal importance of either variable alone. This is precisely the regime where rule mining complements PCA: linear loadings rank individual features, while rule mining exposes the feature \emph{combinations} that jointly characterise each group. The convergence of single-feature and interaction-level evidence on the same compact set of variables indicates that the disparity is driven by a stable, interpretable structure rather than by isolated or run-specific patterns.

\subsection{Mitigation via Fair PCA}

We evaluate whether the disparity identified under our metric can be
mitigated using a Fair PCA correction adapted to the same objective. For
groups $A$ and $B$ and a shared projection $P$, we search for the
rank-$d$ projection minimising
\[
\max\left\{\frac{1}{|A|}\operatorname{loss}(A,P),\;
\frac{1}{|B|}\operatorname{loss}(B,P)\right\},
\]
where $\operatorname{loss}(\cdot,\cdot)$ is the disparity loss of
Definition~\ref{def:loss}, the same quantity used throughout the paper
to discover and characterise the partition. Unlike Samadi et al.'s
original formulation, this objective does not measure raw
reconstruction error but the gap between a group's shared-projection
reconstruction and its own best attainable reconstruction; consequently
the mitigated solution is not required to fall between the two baseline
values, only to minimise the worse of the two.

\paragraph{Global partition:} At $d=9$, the dimension used throughout
our primary analysis, conventional PCA yields a baseline disparity of
$\Delta \approx 5.395$ between the two groups
(loss$_A=0.0624$, loss$_B=5.4576$). Applying our fairness correction
reduces this gap to $\Delta \approx 0.292$
(lossFair$_A=1.7692$, lossFair$_B=1.4773$), a reduction of over 94\%.
Across the full range of dimensions, the corrected loss is not always
bounded between the two baseline values; at several dimensions the
fairness search identifies a valid rank-$d$ projection, confirmed by
near-zero idempotency error, that reduces loss for both groups
simultaneously relative to the naive shared projection.

\begin{figure}[H]
\centering
\includegraphics[width=0.75\textwidth]{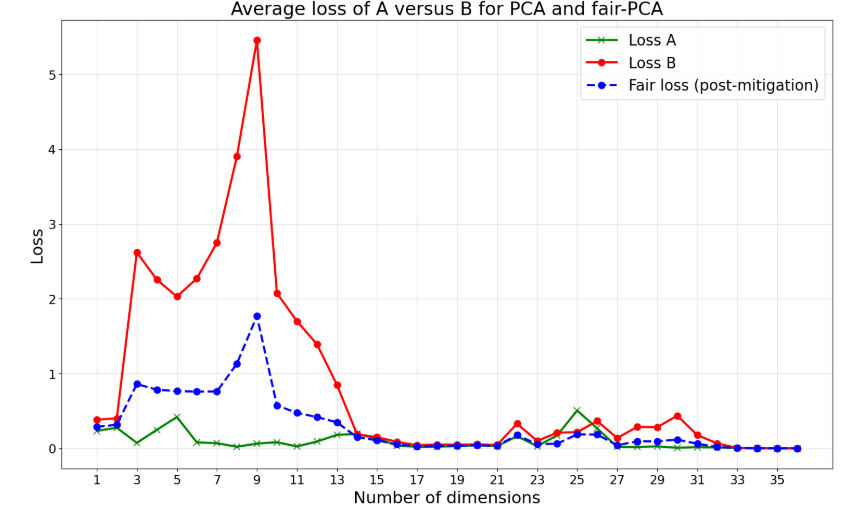}
\caption{
Comparison of normalized subgroup reconstruction losses under
conventional PCA and Fair PCA for the max-disparity partition
identified by the proposed subgroup-discovery procedure.
}
\label{fig:fairpca_global}
\end{figure}

\paragraph{Interpretable well defined attributes:} We repeat this test on two
partitions drawn from the attribution analysis in Section~5. The first
is defined by \texttt{Daytime/\allowbreak evening attendance} (1808
daytime students, 192 evening students). At $d=9$, the baseline
disparity is $\Delta \approx 1.14$ (loss$_A=2.2019$, loss$_B=1.0612$),
reduced to $\Delta \approx 0.0000$ after correction
(lossFair$_A=0.1348$, lossFair$_B=0.1348$); across all sampled
dimensions. Notably, the larger
group bears the higher baseline loss at several dimensions,
underscoring that disparity here is not simply a function of group
size.
\begin{figure}[H]
\centering
\includegraphics[width=0.75\textwidth]{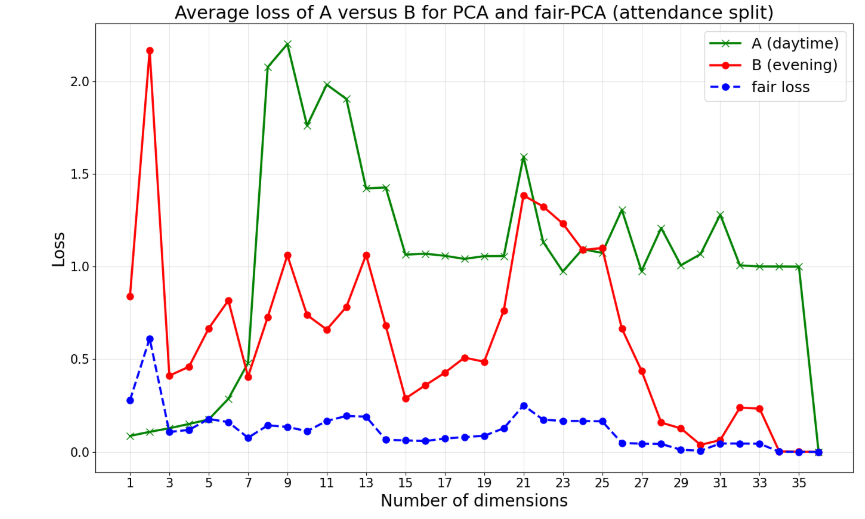}
\caption{
Comparison of normalized subgroup reconstruction losses under
conventional PCA and Fair PCA for the partition defined by
\texttt{Daytime/\allowbreak evening attendance}.
}
\label{fig:fairpca_daytime}
\end{figure}
The second partition is defined by the interaction of
\texttt{Curricular units 2nd sem (grade)} with \texttt{Tuition fees up
to date} (191 students with low grades and unpaid tuition, 1809
otherwise). At $d=9$, the baseline disparity is $\Delta \approx 0.30$
(loss$_A=0.4233$, loss$_B=0.7225$), reduced to $\Delta \approx 0.0000$
after correction (lossFair$_A=0.0926$, lossFair$_B=0.0926$).
\begin{figure}[H]
\centering
\includegraphics[width=0.75\textwidth]{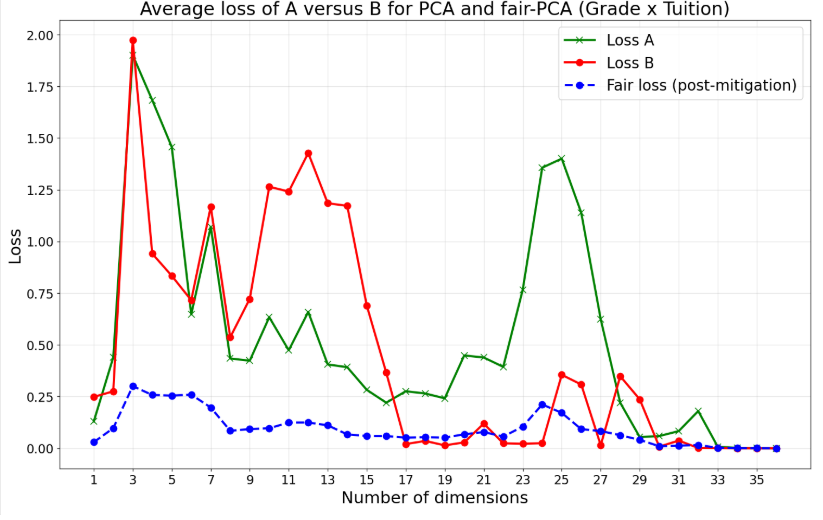}
\caption{
Comparison of normalized subgroup reconstruction losses under
conventional PCA and Fair PCA for the partition defined by the
interaction between \texttt{Curricular units 2nd sem (grade)}
and \texttt{Tuition fees up to date}.
}
\label{fig:fairpca_interaction}
\end{figure}
Across all three settings, the correction
substantially reduces the worse-group loss relative to the naive
shared projection: by approximately 94\% for the globally discovered
partition, and to near-total parity for both externally defined
attribute partitions. The scale of the baseline disparity differs
considerably between settings, largest for the global partition and much smaller for the two interpretable partitions.Nevertheless, in all three cases, however, the correction is
effective regardless of the degree of group-size imbalance.

\subsection{Generalisation on Additional Datasets}
To verify that the behaviour established on the Predict Students' Dropout dataset is not specific to it, we repeat the three-algorithm comparison on two additional datasets drawn from different application domains: the \textit{Default of Credit Card Clients} dataset~\citep{yeh2009comparisons} (23 features, 2{,}000 records, predicting credit-card default), and the \textit{Diabetes Health Indicators} dataset~\citep{teboul2021diabetes} (21 features, 2{,}000 records, predicting diabetes diagnosis from health-survey indicators). For each dataset we apply the same preprocessing pipeline (target encoding of categorical variables, normalisation to $[0,1]$, stratified sampling to 2{,}000 observations) and evaluate Algorithm~1, the two Fixed Projection Sorting variants, and Simulated Annealing under identical hyperparameters. Table~\ref{tab:cross_dataset_results} reports the resulting per-method disparity $\Delta$ together with group sizes.
\begin{table}[H]
\centering
\small
\begin{tabular}{llrrr}
\toprule
\textbf{Dataset} & \textbf{Method} & $|A|$ & $|B|$ & $\Delta$ \\
\midrule
\multirow{4}{*}{Students Dropout}
  & Greedy (Alg.~1)          & 1258 & 742  & \textbf{5.3951} \\
  & FPS - Equal split      & 1000 & 1000 & 0.0966 \\
  & FPS - Matched split    & 1258 & 742  & 0.0301 \\
  & Simulated Annealing      & 1555 & 445  & 4.6591 \\
\midrule
\multirow{4}{*}{Credit Default}
  & Greedy (Alg.~1)          & 406  & 1594 & 2.3063 \\
  & FPS - Equal split      & 1000 & 1000 & 0.0965 \\
  & FPS - Matched split    & 406  & 1594 & 0.6584 \\
  & Simulated Annealing      & 116 & 1884  & \textbf{2.5743} \\
\midrule
\multirow{4}{*}{Diabetes Health}
  & Greedy (Alg.~1)          & 1626 & 374  & 4.8565 \\
  & FPS - Equal split      & 1000 & 1000 & 0.0975 \\
  & FPS - Matched split    & 1626 & 374  & 0.7612 \\
  & Simulated Annealing      & 327  & 1673 & \textbf{5.0940} \\
\bottomrule
\end{tabular}
\caption{Cross-dataset benchmarking of the three partition algorithms. All runs use 2{,}000 observations.}
\label{tab:cross_dataset_results}
\end{table}
A consistent pattern holds across all three datasets: the two methods that co-adapt the partition and the projection:greedy reallocation and simulated annealing attain disparities that exceed both Fixed Projection Sorting variants by a wide margin, typically by one to two orders of magnitude. This separation is the central generalisation result: regardless of domain, jointly optimising the partition together with the subspace recovers far greater representational disparity than any method that scores points under a single fixed projection. The two FPS variants, which hold $P$ fixed at the global PCA solution, remain confined to small disparities on every dataset.

Among the two adaptive methods, greedy and simulated annealing are closely matched, with neither dominating universally. The greedy procedure attains the largest disparity on Students Dropout (with simulated annealing trailing by roughly 22\%), while simulated annealing edges ahead on Credit Default and Diabetes Health (by roughly 10\% and 5\%, respectively). This is the expected relationship between the two: simulated annealing's stochastic exploration is designed to escape the local optima at which a single greedy descent terminates, so it occasionally locates a higher-disparity partition at a substantially greater computational cost. The greedy procedure, by contrast, reaches a high-quality solution in a single warm-started pass, making it the practical default while simulated annealing serves to confirm that the greedy optimum is competitive with a broader global search.


\section{Conclusion}

We have introduced a framework for identifying the binary partition of a
dataset that maximises representational disparity under a shared PCA projection,
without requiring any predefined group labels. The core algorithm adapts the
Fiduccia-Mattheyses local-search paradigm to the reconstruction-loss objective
and is complemented by a fixed-projection sorting baseline and a
simulated-annealing variant; empirically, the greedy procedure attains higher
disparity than both benchmarks. The discovered partition is then characterised
through PCA-based feature scores and association rule mining, which identify
Daytime/evening attendance and Application order as the principal drivers of
representational asymmetry, with Course and Gender emerging as secondary,
interaction-driven contributors within the disadvantaged group. Passing the
discovered partition to Fair PCA achieves loss parity in all tested scenarios,
completing a detect-explain-mitigate pipeline that operates entirely without
predefined sensitive labels.

The finding that representational disparity is driven predominantly by structural
and institutional variables, with demographic attributes playing a secondary role,
is practically significant: it suggests that fairness audits limited to
conventional protected attributes alone may miss a substantial part of the locus
of inequity. Our framework provides a principled mechanism for surfacing such
latent disparities. Two directions stand out for future work. First, the
framework extends naturally to $k \ge 3$ groups, raising algorithmic questions
about the tradeoff between group balance and representational divergence that we
have not addressed here. Second, establishing whether the dominant features are
causally, rather than merely associatively, related to group separation would
sharpen the policy relevance of the attribution analysis.



\appendix
\section*{Appendix}
\addcontentsline{toc}{section}{Appendix}
\section{Full Feature Scores Table (All 36 Variables)}
\label{app:feature-scores}
Table~\ref{tab:feature_scores_appendix} reports the average feature scores across the
full data, the minority group, and the majority group, together with the absolute
loss-difference scores, for all 36 variables. This is the unabridged version of the
ranking summarised in Table~\ref{tab:topfeatures} in the main text.
\begin{table}[H]
\centering
\small
\resizebox{\textwidth}{!}{%
\begin{tabular}{lcccc}
\hline
\textbf{Feature} & \textbf{Total Score} & \textbf{Minority Score} & \textbf{Majority Score} & \textbf{Loss Diff} \\
\hline
Daytime/evening attendance & 37.90\% & 6.86\% & 33.93\% & 27.12\% \\
Application order & 18.83\% & 18.34\% & 17.77\% & 8.47\% \\
Application mode & 6.04\% & 3.73\% & 6.64\% & 3.18\% \\
GDP & 5.18\% & 4.98\% & 5.35\% & 1.80\% \\
Course & 4.52\% & 16.91\% & 4.90\% & 12.60\% \\
Unemployment rate & 4.22\% & 3.76\% & 5.13\% & 2.43\% \\
Tuition fees up to date & 3.84\% & 4.90\% & 4.00\% & 1.50\% \\
Age at enrollment & 3.36\% & 1.93\% & 3.03\% & 1.21\% \\
Debtor & 2.34\% & 3.66\% & 2.93\% & 2.23\% \\
Gender & 1.90\% & 7.25\% & 2.28\% & 5.58\% \\
Inflation rate & 1.58\% & 4.03\% & 1.83\% & 2.80\% \\
Curricular units 2nd sem (grade) & 1.56\% & 2.83\% & 2.13\% & 1.08\% \\
Displaced & 1.55\% & 2.63\% & 2.15\% & 1.41\% \\
Curricular units 1st sem (grade) & 1.38\% & 2.93\% & 2.08\% & 1.11\% \\
Scholarship holder & 1.32\% & 1.17\% & 1.59\% & 0.72\% \\
Marital status & 1.06\% & 0.41\% & 1.00\% & 0.64\% \\
Curricular units 2nd sem (approved) & 0.57\% & 1.49\% & 0.70\% & 0.83\% \\
Curricular units 1st sem (credited) & 0.45\% & 1.96\% & 0.24\% & 1.80\% \\
Curricular units 1st sem (approved) & 0.37\% & 1.75\% & 0.49\% & 1.31\% \\
Curricular units 2nd sem (credited) & 0.33\% & 1.69\% & 0.18\% & 1.58\% \\
Admission grade & 0.28\% & 0.57\% & 0.27\% & 0.44\% \\
International & 0.24\% & 0.37\% & 0.20\% & 0.30\% \\
Curricular units 1st sem (enrolled) & 0.23\% & 1.33\% & 0.15\% & 1.22\% \\
Curricular units 2nd sem (enrolled) & 0.19\% & 0.97\% & 0.15\% & 0.85\% \\
Previous qualification (grade) & 0.14\% & 0.59\% & 0.13\% & 0.48\% \\
Mother's occupation & 0.12\% & 0.16\% & 0.17\% & 0.07\% \\
Curricular units 1st sem (evaluations) & 0.12\% & 0.83\% & 0.11\% & 0.74\% \\
Curricular units 2nd sem (evaluations) & 0.11\% & 1.57\% & 0.19\% & 1.42\% \\
Previous qualification & 0.11\% & 0.06\% & 0.14\% & 0.09\% \\
Mother's qualification & 0.07\% & 0.11\% & 0.05\% & 0.06\% \\
Father's qualification & 0.04\% & 0.07\% & 0.03\% & 0.04\% \\
Educational special needs & 0.03\% & 0.04\% & 0.03\% & 0.03\% \\
Nacionality & 0.01\% & 0.02\% & 0.01\% & 0.01\% \\
Curricular units 2nd sem (without eval.) & 0.01\% & 0.02\% & 0.01\% & 0.02\% \\
Father's occupation & 0.00\% & 0.05\% & 0.01\% & 0.04\% \\
Curricular units 1st sem (without eval.) & 0.00\% & 0.02\% & 0.00\% & 0.02\% \\
\hline
\end{tabular}
}
\caption{Full table of average feature scores across full data, minority groups, majority
groups, and absolute loss-difference scores over 30 iterations of Algorithm~1.
Values for the top variables are summarised in Table~\ref{tab:topfeatures} in the main text.}
\label{tab:feature_scores_appendix}
\end{table}

\section{Background: Association Rule Mining Metrics}
\label{app:arm-background}

For completeness we record the standard definitions of the three metrics used to evaluate association rules~\citep{agrawal1993mining}. For a candidate rule $X \Rightarrow Y$ on a transaction database $\mathcal{D}$:
\begin{itemize}
    \item \textbf{Support}:
    \[
    \text{Support}(X \Rightarrow Y) = \frac{|\{t \in \mathcal{D}: X \cup Y \subseteq t\}|}{|\mathcal{D}|},
    \]
    the proportion of transactions containing both $X$ and $Y$.
    \item \textbf{Confidence}:
    \[
    \text{Confidence}(X \Rightarrow Y) = \frac{\text{Support}(X \cup Y)}{\text{Support}(X)},
    \]
    the conditional probability of observing $Y$ given that $X$ is present.
    \item \textbf{Lift}:
    \[
    \text{Lift}(X \Rightarrow Y) = \frac{\text{Confidence}(X \Rightarrow Y)}{\text{Support}(Y)} = \frac{\text{Support}(X \cup Y)}{\text{Support}(X)\cdot \text{Support}(Y)},
    \]
    the strength of association normalised against statistical independence (a value of
    $1$ indicates independence).
\end{itemize}

\section{Detailed Joint Feature Contributions}
\label{app:interaction-details}

\begin{itemize}
\item \textbf{Curricular units 2nd sem (grade) $\times$ Displaced} (21/30 runs, lift
      1.23--1.54): academic performance and displacement status jointly stratify
      the minority group, the strongest interaction observed among minority-group
      pairings.
\item \textbf{Course $\times$ Curricular units 1st sem (approved)} (21/30 runs, lift
      1.25--1.46): programmatic placement combined with first-semester completion
      characterises the minority group.
\item \textbf{Curricular units 1st sem (grade) $\times$ Displaced} (11/30 runs, lift
      1.23--1.45): first-semester academic performance and displacement status
      jointly distinguish the minority group.
\item \textbf{Curricular units 2nd sem (approved) $\times$ Tuition fees up to date}
      (21/30 runs, lift 1.36--1.62): course completion combined with tuition status
      is the strongest interaction observed among majority-group pairings.
\item \textbf{Curricular units 2nd sem (grade) $\times$ Tuition fees up to date}
      (22/30 runs, lift 1.33--1.61): academic performance combined with tuition
      status characterises the majority group.
\item \textbf{Curricular units 1st sem (approved) $\times$ Tuition fees up to date}
      (21/30 runs, lift 1.34--1.61): first-semester course completion combined with
      tuition status reinforces majority-group separation.
\end{itemize}

\bibliographystyle{abbrvnat}
\bibliography{References}

\end{document}